\UseRawInputEncoding
\documentclass{article} 
\usepackage{iclr2022_conference,times}

\usepackage{amsmath,amsfonts,bm}

\def\eqref#1{equation~\ref{#1}}

\def\1{\bm{1}}

\DeclareMathAlphabet{\mathsfit}{\encodingdefault}{\sfdefault}{m}{sl}
\SetMathAlphabet{\mathsfit}{bold}{\encodingdefault}{\sfdefault}{bx}{n}

\usepackage{hyperref}
\usepackage{url}
\usepackage{times}
\usepackage{helvet}
\usepackage{courier}
\usepackage{CJKutf8}
\usepackage{tabularx}
\usepackage{svg}
\usepackage{multirow}
\usepackage{graphicx}
\usepackage{adjustbox}
\usepackage{cite}
\usepackage{algorithm}
\usepackage{algpseudocode}
\usepackage{makecell}
\usepackage{float}

\title{Green CWS: Extreme Distillation and Efficient Decode Method Towards Industrial Application}

\author{
\centerline{Yulan Hu\textsuperscript{1,2}, Yong Liu\textsuperscript{1}} \\
\centerline{\textsuperscript{1}Gaoling School of Artificial Intelligence, Renmin University of China, Beijing, China} \\
\centerline{\textsuperscript{2}Kuaishou MMU} \\
\centerline{\texttt{\{huyulan, liuyonggsai\}@ruc.edu.cn}} \\
}

\begin{document}

\maketitle

\begin{abstract}
Benefiting from the strong ability of the pre-trained model, the research on Chinese Word Segmentation (CWS) has made great progress in recent years. However, due to massive computation, large and complex models are incapable of empowering their ability for industrial use. On the other hand, for low-resource scenarios, the prevalent decode method, such as Conditional Random Field (CRF), fails to exploit the full information of the training data. This work proposes a fast and accurate CWS framework that incorporates a light-weighted model and an upgraded decode method (PCRF) towards industrially low-resource CWS scenarios. First, we distill a Transformer-based student model as an encoder, which not only accelerates the inference speed but also combines open knowledge and domain-specific knowledge. Second, the perplexity score to evaluate the language model is fused into the CRF module to better identify the word boundaries.  Experiments show that our work obtains relatively high performance on multiple datasets with as low as 14\% of time consumption compared with the original BERT-based model. Moreover, under the low-resource setting, we get superior results in comparison with the traditional decoding methods.
\end{abstract}

\section{Introduction}
Chinese Word Segmentation (CWS) is typically regarded as a fundamental and preliminary step for NLP tasks, and it has attracted heavy studies these years. Normally CWS is considered as a sequence labeling problem. That is, each character in a sentence was labeled by symbols among 'B, M, E, S,' which indicates the beginning, middle, and end of a word, or a word with a single character. 

In the past few years, methods have been proposed to improve the performance of CWS. \citet{huang2015bidirectional} use bidirectional LSTM to capture the long dependency features of sequence, and the proposed BiLSTM-CRF became a strong baseline for the sequence tagging task. \citet{yang2018subword} apply lattice LSTM network to integrate character information with subword information, which significantly enhances the performance. \citet{tian2020improving} propose a neural framework, WMSEG, which uses a memory network to incorporate wordhood information as strong contextual features for the encoder. 

In addition to designing the ancillary modules for the CWS task, following the development of pre-trained models \citep{han2021pre}, CWS achieves new progress. \citet{huang2019toward} fine-tune BERT for multi-criteria CWS, where a private projection layer is adopted to extract criteria-specific knowledge. \citet{ke2020pre} present a CWS-specific pre-trained model, which employs a unified architecture to make use of segmentation knowledge of different criteria. \citet{liu2021lexicon} raise a lexicon enhanced BERT, which combines the character and lexicon features as the input. Besides, it attaches a lexicon adapter between the Transformer layers to integrate lexicon knowledge into BERT.
Though pre-trained models significantly improve the overall performance of CWS, the practicality is not that satisfactory. Under industrial scenarios that require a quick response, the whole system can only afford a few milliseconds for CWS, hence the model structure is dedicate to be simple and concise to save the inference time. \citet{duan2019attention} come up with a Transformer variant, Gaussian-masked Directional (GD) Transformer, to take the unigram features for scoring and using a bi-affine attention scorer to directly predict the word boundaries. The proposed method gains high speed with comparable performance. \citet{huang2019toward} utilize knowledge distillation to get a shallow model. Meanwhile, the quantization methods are investigated for network acceleration. 

Concerning CWS towards low-resource scenarios in industrial, for example, in Table 1, the original query comes from a famous Chinese video-sharing social network. In the absence of external knowledge or strategy, it is challenging to segment the query correctly to recall what we want. The BiLSTM-based model segments the sentence correctly while fails to satisfy the industrial speed requirement. We extract test sentences from different corpora, assuming that the time left for the segmentation for a sentence is 1ms, on account of the average length is 35, then the minimum speed is around 68KB/s. Obviously, the CNN-based and BiLSTM-based models both fail to meet the threshold.
\begin{table*}[h]
\begin{center}
\begin{tabular}{|c|c|c|}
\hline
\textbf{Tools} & \textbf{Segmentation Results} & \textbf{Speed}  \\
\hline
\textbf{Baseline} & \makecell[c]{\begin{CJK}{UTF8}{gbsn} 新来\; 的\; 吃鸡 \; 主播\end{CJK}\\ The newcome chiji host} & 68KB/s \\ \hline
Jieba & \makecell[c]{\begin{CJK}{UTF8}{gbsn}新\; 来\; 的\; 吃\; 鸡主播\end{CJK}} & 673KB/s\\ \hline
Thulac & \makecell[c]{\begin{CJK}{UTF8}{gbsn} 新\; 来\; 的\; 吃\; 鸡\; 主播 \end{CJK}} & 103KB/s \\ \hline
PKUSeg & \makecell[c]{\begin{CJK}{UTF8}{gbsn} 新来\; 的\; 吃\; 鸡\; 主播 \end{CJK}} & 123KB/s\\ \hline
CNN-based model & \makecell[c]{\begin{CJK}{UTF8}{gbsn} 新\; 来\; 的\; 吃\; 鸡\; 主播 \end{CJK}} & 36KB/s\\ \hline
BiLSTM-based model & \makecell[c]{\begin{CJK}{UTF8}{gbsn} 新来\; 的\; 吃鸡\; 主播 \end{CJK}} & 12KB/s\\ \hline
\end{tabular}
\caption{Illustration of different segmentation results, \textbf{chiji} is the alias of a famous mobile game, the \textbf{Baseline} row gives the golden segmentation result, the minimum speed need is 68KB/s.}
\end{center}
\end{table*}

So the problem arises, given a light-weighted CWS structure, how to improve the overall performance as much as possible? There are two major methods corresponding to different phases to alleviate the problem. During the encoding phase, we can make full use of dictionary knowledge. \citet{liu2019neural} adopt two methods, named Pseudo Labeled Data Generation and Multi-task Learning to make full use of dictionary knowledge to promote the performance of CWS. Also, it is classical to utilize transfer learning technology, \citet{xu2017transfer} initialize a student model with the learned knowledge from the teacher model, and a weighted data similarity method is proposed to train the student model on low-resource data. Besides, we can ease the problem during the decoding process by modifying the idiomatic decode paradigm, however, to our best knowledge, there is not much research on this part.

We optimize our work from two aspects to build a fast and accurate CWS segmenter towards the industrial context. First, we utilize the knowledge distillation to accelerate the model, and shallow Transformer-based student models which is distilled as the encoder. Second, we believe that the sequence-level semantic information is helpful during the decoding process. We propose a novel CRF, named PCRF, which merges the sequence-level perplexity score into the original CRF, enhances the CWS performance in the low-resource scene. Lastly, the segmentation results are post-processed based on a coarse-grained dictionary extracted from the training corpus.


Overall, the main contributions of this work can be summarized as follows:
\begin{itemize}
\item Model accelerating technique is realized to build a fast segmenter for industrial application. The light-weighted structure not only achieve the speed request towards real-time scenarios but also combines open knowledge from the teacher and in-domain knowledge from the training corpus.

\item An upgraded CRF named PCRF is proposed. Apart from the character-level knowledge learned by the encoder, the sequence-level knowledge is acquired through a simple n-gram model. The two kinds of knowledge are jointly sent to the PCRF for decoding.

\item We create two CWS datasets. The original corpus is extracted from e-commerce and video sharing social network. The two CWS datasets are cross annotated carefully by three experts in Chinese linguistics. We plan to release these two datasets gradually for further academic research.

\item Experiments on more than ten corpora, include eight benchmark datasts, two publicly available datasets, and two hand-crafted datasets, have demonstrated the effectiveness of our model. We believe that the proposed CWS framework is valuable for the industrial scenario, especially in the areas that lack labeling data and demand high inference speed.
\end{itemize}

\section{Background}
In this section, we will first describe the problems of the CWS model in industrial and then introduce the main concepts related to the proposed work.
\subsection{Task Difficulty}
Towards CWS in the actual scenario, there are three problems to deal with. Firstly, as new media burst endlessly, sending annotators to tag the massive unlabeled corpus is labor-wasting. Meanwhile, the distribution of the publicly available datasets is very different from the industrial data, though there may be common knowledge that we can make use of. Secondly, it is easy to train a high-performance BERT-based CWS model and deploy it in the system. However, given enough acceleration techniques, the large model still fails to meet the milliseconds or even tenths of a millisecond speed needs in the rigorous environment. Thirdly, the publicly available CWS tools are dissatisfied, especially in new areas, which brings much extra work like dictionary mining, semantic disambiguation, etc.

\subsection{Feature Extraction}
Feature extraction is the necessary step for CWS task, through feature extraction, the contextual knowledge can be refined to ensure the label of each character. By utilizing the fine-tune scheme of the pre-trained models, the characters are first mapped into embedding vectors and then fed into several encoder blocks, given a sentence $X=\left\{x_{1}, x_{2}, \ldots, x_{n}\right\}$, the feature can be represented as:
$$\mathbf{h}_{i}=\operatorname{BERT}\left(\mathbf{e}_{1}, \mathbf{e}_{2}, \ldots, \mathbf{e}_{n} ; \theta\right),$$
where $\theta$ denotes the parameters learned during the fine-tuning phase. 

\subsection{Inference Layer}
After feature extraction, we will get the scores of each character corresponding to the tag set, specifically the transition score and emission score. The scores will be sent to CRF or MLP decoder to get the best tagging result.
Formally, the probability of a label sequence is formalized as:
$$p(Y \mid X)=\frac{\Psi(Y \mid X)}{\sum_{Y^{\prime} \in \mathcal{L}^{n}} \Psi\left(Y^{\prime} \mid X\right)},$$
where $\Psi(Y \mid X)$ is the potential function:
$$\Psi(Y \mid X)=\prod_{i=2}^{n} \psi\left(X, i, y_{i-1}, y_{i}\right),$$
$$\psi\left(X, i, y_{i-1}, y_{i}\right)=\exp \left(s(X, i)_{y_{i}}+\mathbf{b}_{y_{i-1} y_{i}}\right),$$
where $y$ denotes the tag label, $\mathbf{b} \in \mathbf{R}^{|\mathcal{L}| \times|\mathcal{L}|}$ is a trainable parameter denotes the transition score between the tag set, which can be randomly initialized. $\mathbf{b}_{y_{i-1} y_{i}}$ means the score from $\mathbf{y_{i-1}}$ to $\mathbf{y_{i}}$, $s(X, i)_{y_{i}}$ denotes the emission score respective to each tag label for $\mathbf{i_{th}}$ character:
$$s(X, i)=\mathbf{W}_{s}^{T} \mathbf{h}_{\text {i}}+\mathbf{b}_{s},$$
$\mathbf{W}_{s}^{T} \in \mathbf{R}^{|\mathcal{L}| \times|\mathcal{L}|}$ and $\mathbf{b} \in \mathbf{R}^{|\mathcal{L}|}$ are trainable parameters.
If using MLP decoder, the output label can be easily acquired through the Softmax function.

\subsection{Perplexity Score}
The perplexity score (PPL), usually used as the evaluation metric for language model, can depict how reasonable a sentence is. Given a segmented sentence $S$, the PPL of $S$ can be calculated as:
$$\begin{array}{l}
PPL(S)=p\left(w_{1}, w_{2}, w_{i}, \ldots, w_{m}\right)^{-1 / m}\quad=\sqrt[m]{\prod_{i=1}^{m} \frac{1}{p\left(w_{i} \mid w_{1}, w_{2}, \ldots, w_{i-1}\right)}},
\end{array}$$
where $w_i$ denotes the word in the sentence, take bi-gram language model as an instance, under the Markov hypothesis, the current word merely relies on the former word in the sentence, so the calculation can be reduced to:
$$PPL(S)\approx p(w_{1})p(w_{2}\mid w_{1})p(w_{3}\mid w_{2})..p(w_{n}\mid w_{n-1}).$$

\section{Model Description}
In this work, we propose a hierarchical CWS model towards real industrial scenarios usage. The model contains two essential parts, a distilled Transformer-based encoder and a concisely upgraded CRF decoding layer (PCRF). We will describe the two parts respectively.
\subsection{Distilled Encoder}
To balance the accuracy and computational cost, firstly, we fine-tune the pre-trained model on the CWS training data to get a relatively high-performance teacher model. Then, we distill knowledge from the teacher model into a smaller Transformer-based student network. Transformer is recognized to have an advantage over RNN and CNN in feature extraction.
During the distillation process, a cross-entropy ($\mathcal{CE}$) function is adopted to calculate the divergence between the output logits from the teacher model and student model. The overall loss function, incorporating both distillation and student losses, is calculated as:
$$\mathcal{L}(x ; \theta_s)=\alpha * \mathcal{CE}\left(y, p\right)+\beta * \mathcal{CE}\left(\sigma\left(l_{t} ; T\right) \sigma\left(l_{s}, T\right)\right),$$
where $x$ is the input data, $p$ is the hard label predicted by the student model, $\theta_s$ denotes the student model parameters, $l_s$ and $l_t$ are the logits of the student and teacher, respectively, $\sigma$ is the softmax function parameterized by the temperature $T$, $\alpha$ and $\beta$ are coefficients to trade off the two losses.
\subsection{PCRF Inference Layer}
The conventional CRF is usually served as the decoding component after the neural encoder. By aggregating the emission score and the transition score, the CRF decode module assigns a score for each possible sequence of tags through the Viterbi algorithm.

However, it can be troublesome if lacks of sufficient training corpus. In industry, there is often a need for CWS in new scenarios. Consider that the emission score actually reflects the capability of the prepositive encoder, which is, the encoder can not perfectly generalize its ability to unseen words (OOV); thus, the emission score can be biased. In other words, as clearly analyzed in \citet{wei2021masked}, the author states that without a hard mechanism to enforce the transition rule, the conventional CRF can result in the occasional occurrence of illegal predictions, which indicates that it is possible to lead to wrong tag paths under the current decode framework.  

To this end, we believe that it relies solely on the emission score and the transition score are insufficient, especially in low-resource circumstances. Before that, it was common to segment text based on a language model, while to our best knowledge, we did not find works to incorporate the language model with the decoding process. Our solution is intuitive. Apart from the already acquired two scores for each character, we will calculate the PPL of the already decoded sequence. For example, as illustrates in Figure 1, the sentence \begin{CJK}{UTF8}{gbsn}请播放一首将军令\end{CJK} means "Please play 
General's order". At the index of character \textbf{\begin{CJK}{UTF8}{gbsn}军\end{CJK}}, one possible decoded path is \begin{CJK}{UTF8}{gbsn}(请 \ 播放 \ 一首 \ 将 \ 军)\end{CJK}, another is \begin{CJK}{UTF8}{gbsn}(请 \ 播放 \ 一首 \ 将军)\end{CJK}, the corresponding decoding score at the character \begin{CJK}{UTF8}{gbsn}军\end{CJK} can be formulated as:

\begin{center}
\begin{equation}
Score_1 = E_I+T_{BI}+\lambda * PPL(Path_1)
\end{equation}
\end{center}

\begin{center}
\begin{equation}
Score_2 = E_B+T_{BB}+\lambda * PPL(Path_2)
\end{equation}
\end{center}


where $E$ is emission score, $T$ is the transition score,  $\lambda$ is the coefficient to measure the importance of PPL, which ranges between zero to one. 

\begin{figure}[h]
\caption{An decode example of original CRF VS PCRF, at the index of character \begin{CJK}{UTF8}{gbsn}军\end{CJK}, the emission score for the label "I" is smaller than "B", unfortunately, the CRF module fail to rectify the illegal path after the dynamic decoding process ended. In contrast, the perplexity score calculated by the language model, together with the aforementioned two scores, managed to justify that the character \begin{CJK}{UTF8}{gbsn}军\end{CJK} should be the middle of a word rather than the start of a word.} 
\label{fig}
\end{figure}

Note that there is a key point, the n-gram language model is derived from a large labeled training corpus that is not realistic in industrial. There are solutions to deal with such a chicken-and-egg problem. On one hand, it is acceptable that we can merely depend on the training corpus as long as the training data is abundant. On the other hand, the main difference between various CWS corpora lies in the tagging criteria. Through the investigation in \citep{fu2020chinese}, the distance between different datasets can be quantitatively characterized, we adopt the label consistency of words that appear in the training set as the distance measure metric, defined as:
$$
\psi\left(w_{i}^{l}, D^{train}\right)=\left\{\begin{array}{ll}
0 & \left|w_{i}^{train}\right|=0 \\
\frac{\left|w_{i}^{train, l}\right|}{\left|w_{i}^{train}\right|} & \text { otherwise },
\end{array}\right.
$$
where $\left|w_{i}^{train, l}\right|$ represents the occurrence of word $w_{i}$ with label $l$ in the training set, $D^{train}$ is the training set. Naturally, we can replenish the current dataset with available datasets that are adjacent to it. Unlike the conversational way, we use the adscititious datasets to train the n-gram language model to fuse the knowledge into the decoding process.



Based on the above discussion, we propose the PCRF decode framework, formally described in Algorithm 1.


\begin{algorithm}
\caption{PCRF}\label{alg:cap}
\begin{algorithmic}[1]
\Require tag set $L$, Emission score $E=e_{it}$, Transition score $T=t_{ij}$, PPL coefficient $\lambda$.
\Ensure Empty array $Score$, Empty array $Path$, length of the sentence $L_S$.

\While{$L_S$ is not met}
\State Add $E_{it}$ and $T_{ij}$
\For{$t\in L$}
\State $Get \ Corresponding \ PPL_t $
\State $Score_{it}$ $\gets$ $E_{ij}$ $+$ $T_{ij}$ $+$ $\lambda$ $*$ $PPL_i$
\EndFor
\State $Update \ Score \ and \ Path$
\EndWhile
\end{algorithmic}
\end{algorithm}


\section{Experiments}
\textbf{Datasets} We collected eight standard benchmark CWS datasets, include MSR, AS, PKU and CityU from SIGHAN2005, UDC from CoNLL 2017 Shared Task \citep{zeman2018conll}, SXU from SIGHAN2008, CTB from \citep{xue2005penn} and CNC corpus. Also, two publicly available datasets are adopted, 
include WTB from \citep{wang2014dependency} and ZX from \citep{zhang2014type}, the WTB is collected from Sina Weibo, a popular social media site in China. The ZX dataset is derived from Zhuxian, a fairy tale novel with a literature genre similar to martial arts novels. Besides, we develop two self-built datasets, the ECOMM, and CAPTION. The original corpus of ECOMM comes from the caption of commodities on the e-commerce website, which depicts the features of the commodities, and we collect CAPTION dataset from a video sharing social website, mainly the search queries of the short videos. Three linguistics experts are invited to annotate the two datasets. When disagreement with the annotation results occurs, they will discuss and align with each other so that will be no further conflict. The two datasets will be released gradually in the future. The WTB, ZX, ECOMM, CAPTION datasets are denoted as low-resource datasets in this work. Details of the datasets after preprocessing are shown in Table 2.

\begin{table*}[]
\centering
\begin{tabular}{c|cc|cc|cc}
\hline
Datasets & \multicolumn{2}{|c}{Words} & \multicolumn{2}{|c|}{Phrases}  &\multicolumn{2}{|c}{ASL} \\ \hline
&Train & Test &  Train &Test&  Train &Test   \\ \hline
MSR & 2223k & 252k & 50k & 8k & 46.59 & 46.58 \\
AS & 5461k & 123k & 70k & 4k & 11.81 & 11.14 \\ 
PKU &1121k &104k & 21k & 3k & 95.92 & 88.87 \\
CityU &1400k &40k & 30k & 2k & 45.31 & 45.38 \\
CTB & 730k & 53k & 17k & 2k & 45.71 & 43.93 \\
UDC & 111k & 12k & 3k & 0.6k & 39.22 & 38.42 \\
CNC & 6569k & 726k & 52k & 14k & 43.28 & 43.06 \\
SXU & 540k & 113k & 12k & 3k & 49.85 & 50.43 \\ \hline
WTB & 16k & 0.2k & 0.6k & 0.1k & 28.71 & 32.97 \\
ZX & 88k & 33k & 0.3k & 0.7k & 39.62 & 34.39  \\
ECOMM & 29k & 3k & 2k & 0.3k & 34.26 & 34.96  \\
CAPTION & 215k & 24k & 20k & 3k & 9.69 & 9.82  \\ \hline
\end{tabular}
\caption{Details of the twelve datasets. Three aspects of information are exhibited. \textbf{Words} represent the number of tokens that appear in the dataset. \textbf{Phrases} represent the number of tokens in the dataset whose length is greater than two. \textbf{ASL} is the abbreviation of "average sentence length," which demonstrates the average length of sentences in a dataset.} 
\end{table*}

\textbf{Hyper-parameters} We strictly follow the steps of knowledge distillation. Specifically, we adopt pre-trained BERT \citep{devlin2018bert} as the teacher model, three Transformer-based student models with layers 1, 3, and 6 are distilled. We use Adam with the learning rate of 5e-5, the batch size is set to 8, the max length is set to 512, the temperature for distillation is set to 8. We select the last saved checkpoint (the 20th epoch) as the student model. And the F1 score is used to evaluate our model.

\textbf{Language Model}
We adopt KenLM \citet{heafield2011kenlm}, a fast and simple language model toolkit to calculate the PPL score. KenLM provides an intuitive API to train n-gram language models as well as inferencing.

\textbf{Preprocessing}
Among all datasets, extra space between tokens and invalid characters is removed. We transform AS and CityU from the traditional form into a simplified form. Besides, all tokens are converted into half-width.


\subsection{Overall Results}
In this section, We first give the main results on the four low-resource datasets. Then we crop the benchmark training set to a different scale while keeping the test set still to simulate the low-resource scenarios. For simplicity, we distill a single-layer Transformer as the base encoder.

\textbf{Low-resource datasets} The pre-trained language models have shown us its magic power. By fine-tuning the model, it's not challenging to get a satisfying performance on CWS. We do not intend to prove the superiority of the pre-trained models in the CWS task. From the perspective of effectiveness and practicality,  we choose two kinds of comparison models. The first is the CWS research works towards low-resource or high-speed needs scenarios, and the second is the widely used neural network-based CWS tools. The experimental results are shown in Table 3.
 

 
Specifically, the truncated (1, 3, or 6 layers) BERT learned from the teacher are used as the encoder, "KD\_1" stands for knowledge distillation with single Transformer layer in Table 3. We use Softmax, CRF, and the proposed decode method (PCRF) to implement the decoding process. Overall, eliminating the contrast models, our model achieves nine best performances among three group experiments (twelve in total), sufficient to proof the effectiveness of our model especially when the model is shallow. Under the single-layer setting, our model outperforms the CRF-based model by 1.43\%, 3.10\%, 1.80\%, and 5.65\%, respectively, on four datasets. As the model gets deeper, the results are still convictive. Besides, we noticed that in the ECOMM dataset, the results are not as persuasive as the rest datasets. We believe the reason could be the distribution of the ECOMM dataset is far different from the rest datasets.
The corpus is more like a vocabulary stack rather than fluent sentences, so the PCRF modules conversely hurt performance when the model gets deeper.

\begin{table*}[t]
    \centering
    \begin{tabular}{ccccccc}
        \hline
        Models & ZX & WTB & ECOMM & CAPTION \\ \hline
        Jieba  & 79.45 & 81.23 & 86.70 & 74.96  \\
        PKUSeg  & 87.45 & 87.24  & 83.51 & 71.40  \\
        THULac  & 83.28 & 82.33 & 69.35 & 75.59 \\
        \citet{huang2019toward} & 97.0 & 93.1 & - & - \\
        \citet{huang2020joint} & 96.77 & - & - & - \\
        \hline\hline
        KD\_1+Softmax & 92.57 & 82.46 & 88.08 & 75.93 \\
        KD\_1+CRF & 92.57 & 82.55 & 88.12 & 75.95 \\
        Ours(KD\_1+PCRF) & \textbf{94.00} & \textbf{85.65}  & \textbf{89.92} & \textbf{81.60}\\

        \hline\hline
        KD\_3+Softmax   & 96.07 & 90.17 &94.61& 82.10   \\
        KD\_3+CRF  & 96.07 & 90.17 &\textbf{94.66}& 82.08  \\
        Ours(KD\_3+PCRF) &\textbf{96.12}&\textbf{90.45}&94.16& \textbf{83.00}   \\
        \hline\hline      
        KD\_6+Softmax &\textbf{96.74} & 91.94 &95.17 & 83.17  \\
        KD\_6+CRF & 96.73 & 91.94 & \textbf{95.22} &  83.20 \\
        Ours(KD\_6+PCRF) & 96.71 & \textbf{92.38} & 94.65& \textbf{83.31} \\ 
        \hline\hline
        Ours(Teacher\_12) & 96.69 & 92.43 & 95.77& 83.79 \\ 
        \hline\hline
    \end{tabular}
    \caption{Results on four low-resource datasets.}
\end{table*}

\textbf{Low-resource settings} The analysis of the former part proves the usefulness of the proposed model. Nonetheless, given sufficient training data and deeper layers, the results are closer than the results in low-resource and shallow scenarios. The average divergence between the KD\_1+CRF and our model (KD\_1+PCRF) of four in-domain datasets is 11.98\%, while 0.96\% for the six layers condition, which means that the proposed method is more robust in the low-resource scenario and shallow layers. 

To further evaluate the model towards the low-resource scenario, we invariably keep the validation set and test set and randomly choose a different ratio of the universal datasets as the new training data, which simulate the extreme low-resource scenario. The ratio is set to 50\%, 30\%, 10\%, and 1\% descendingly. The smaller the number, the closer to the setting of the low-resource scenario. We adopt the KD\_1 as the base encoder. Table 4 shows that our model achieves advantages in most experiments (30 out of 33), with 1.29\% improvement on average under 50\% training data, 2.41\% improvement on average under 30\% training data, 5.07\% improvement on average under 10\% training data, and 16.60\% improvement on average under 1\% training data. Meanwhile, we find that the three experiments with reduced performance all occurred on the AS dataset. Notice from Table 2 that the average length of the AS dataset is relatively short, as will be discussed in the next section, the PCRF-based model is more efficient for long sentences.

\begin{table*}[h]
\begin{tabular}{cccccccccl}
\hline
\multicolumn{1}{l}{RATIO} & Models         & MSR       & AS       & PKU  & CityU  & CTB    & SXU      & UDC    &  CNC    \\ \hline

\multirow{3}{*}{50\%}       & KD\_1+Softmax     & 93.79  &  \textbf{95.18}      &  89.75   & 89.87 & 90.56 &  89.69  & 81.68 &  95.02 \\ \cline{2-10} 
                            & KD\_1+CRF         & 93.79  &  95.18      &  89.75   & 89.88 & 90.56 &  89.70  & 81.69 &  95.02 \\ \cline{2-10} 
                            & Ours(KD\_1+PCRF)             & \textbf{94.28}    &  94.11  &  \textbf{91.93} & \textbf{90.46} & \textbf{91.73} & 
                            \textbf{91.98}  & \textbf{86.12} &    \textbf{95.30}\\  \hline\hline
                            
\multirow{3}{*}{30\%}       & KD\_1+Softmax    & 92.34    &   \textbf{94.78}    &  87.78   & 88.29 & 89.16 &  87.96  & 78.18 &  94.22  \\ \cline{2-10} 
                            & KD\_1+CRF        & 92.34    &   94.78    &  87.80   & 88.29 & 89.16 &  87.96  & 75.46 & 94.22 \\ \cline{2-10} 
                            & Ours(KD\_1+PCRF)     & \textbf{93.84}    &     93.92     &  \textbf{90.74}   & \textbf{89.95} & \textbf{90.80} &  \textbf{91.03}  & \textbf{84.07} &   \textbf{94.93} \\ \hline\hline                            
                            
\multirow{3}{*}{10\%}        & KD\_1+Softmax   & 88.80   &  \textbf{93.37}  &  83.53   & 84.39 & 84.66 &  83.09  & 64.73 &  91.90 \\ \cline{2-10} 
                            & KD\_1+CRF        & 88.81   &  93.36  &  83.55   & 84.39 & 84.67 &  83.10  & 64.73 &   91.90  \\ \cline{2-10} 
                            & Ours(KD\_1+PCRF)     & \textbf{92.54}    &  93.33  &  \textbf{87.90}   & \textbf{87.36} & \textbf{89.19} &  \textbf{88.58}
                            & \textbf{82.18} &   \textbf{93.96}  \\ \hline\hline
                            
\multirow{3}{*}{1\%}        & KD\_1+Softmax          & 79.26    &   86.27     &  55.55   & 66.69 & 55.66 & 52.61  & 41.12 &  82.77\\ \cline{2-10} 
                            & KD\_1+CRF        & 79.27    &  86.27  &  55.80   & 66.69 & 55.79 &  52.57  & 41.18 &    82.77 \\ \cline{2-10} 
                            & Ours(KD\_1+PCRF)     & \textbf{89.14}    &  \textbf{90.46}  &  \textbf{77.34}   & \textbf{80.16} & \textbf{78.32} &  \textbf{84.84} 
                            & \textbf{65.93} &  \textbf{89.16} \\ \hline                            
                            
\end{tabular}
\caption{Results on eight universal dataset results at different scales.}
\end{table*}

\subsection{Scalability}
As mentioned above, unlike some scenarios of competing performance, industrial scenarios represented by searching, advertising, and recommendation, the speed of CWS is critical as performance. According to the experiment, the KD\_1 based CWS model spends roughly 14\% of time consumption compared with the original BERT model, which meets the requirement of segmentation response. 

Since we adopt the proposed PCRF as the decoding module, it relies highly on the language model. Intuitively the length of a sentence can have a substantial effect on the decoding performance. We randomly select several test sets and classify the sentence in each test set according to its length. We denote that the length smaller than 20 as short sentence, the length longer than 50 as the long sentence, otherwise the medium sentence. Note that the numbers of the long and short test sets are inconsistent, so we randomly select the size of the smaller dataset from the larger data set to make the equal size of the two datasets. To directly exhibit the influence of the length of the sentence and avoid contingency, we use three distilled student models of diverse layers (1, 3, and 6) as the encoder and conduct experiments on the short and long test groups. Table 5 shows the result divergence between the two test groups, the results are averaged by the number of student models. Given that the sentence length is the only variable, we can conclude that the proposed decode method will benefit when the sentence length is long.
\begin{figure}[htb]
\centerline{\includegraphics[width=0.8\textwidth]{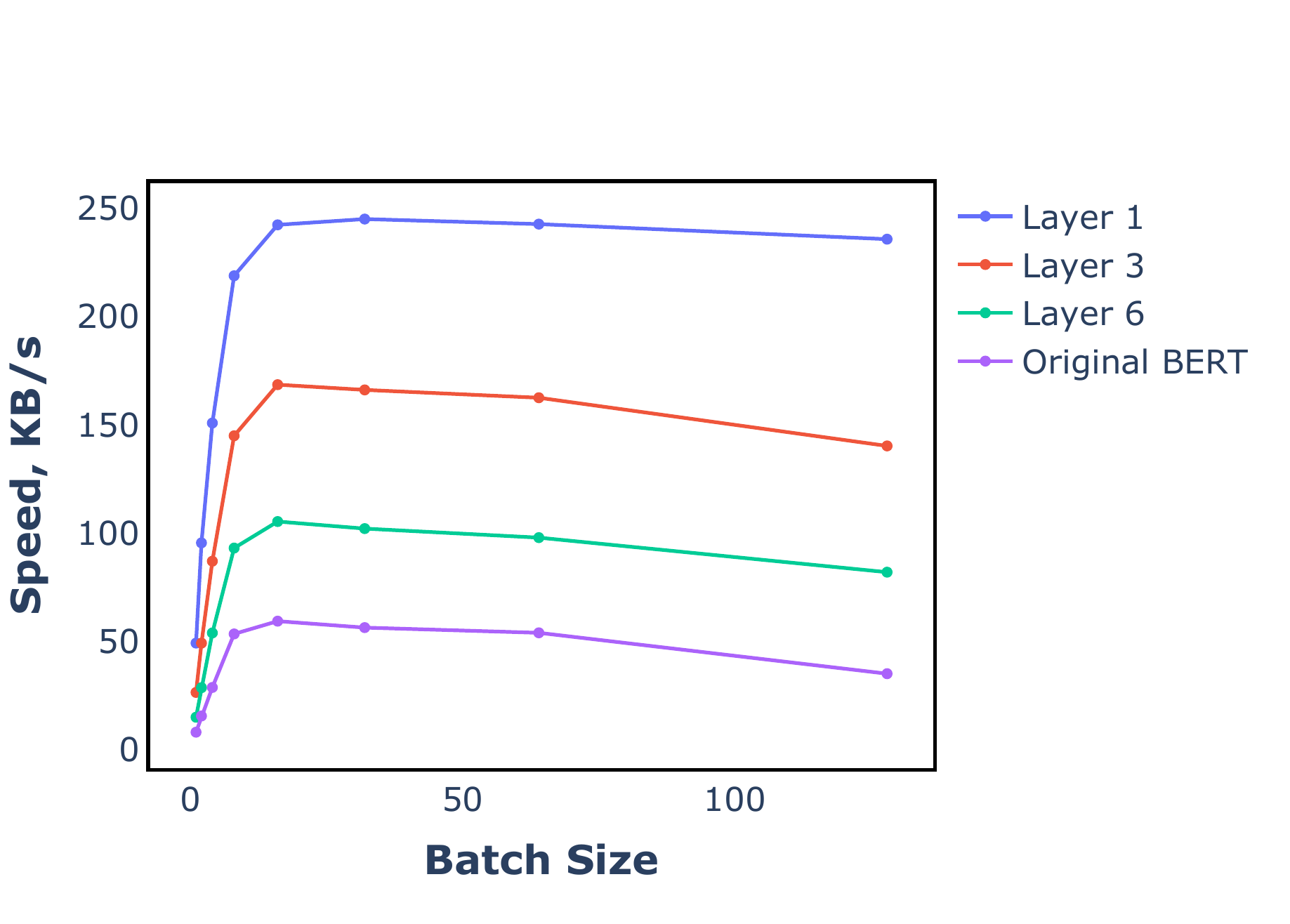}}
\caption{Inference speed at different batch size, the sequence length is set to 128.}
\label{fig}
\end{figure}

\begin{table*}[h]
\begin{tabular}{ccccccccl}
\hline
  
\multicolumn{1}{l}{Models} & Datasets         & CityU  & PKU   & CTB   & MSR    & SXU  & UDC &CNC  \\ \hline

\multirow{2}{*}{\textbf{CRF Based}}     & Short   & 0.7804  &  0.7465  &  \textbf{0.8153}   & 0.9270 & \textbf{0.7878} & \textbf{0.6822}  & \textbf{0.9151} \\ \cline{2-9}
                                        & Long    & \textbf{0.8236}  &  \textbf{0.7961}  &  0.7977   & 0.9270 & 0.7859 & 0.6430  & 0.9090  \\ \hline\hline        
                                    
\multirow{2}{*}{\textbf{PCRF Based}}    & Short    & 0.8346  &  0.8236  &  0.8675   & 0.9080 & 0.8761 & 0.7661 & \textbf{0.9364}  \\ \cline{2-9}
                                        & Long     & \textbf{0.8739}  &  \textbf{0.8708}  &  \textbf{0.8746}   & \textbf{0.9270} & \textbf{0.8878} & \textbf{0.7766} & 0.9330 \\ \hline\hline                                     
                                    
\end{tabular}
\caption{Results on eight universal dataset results at different scales.}
\end{table*}

Figure 2 illustrates the inference speed under various batch sizes. We distilled student models with 1, 3, and 6 layers, respectively. With the increase of batch size, the inference speed of each model will increase simultaneously. As the batch size comes to a certain extent, the speed reaches a maximum point and stop increases any more, and even decreases. Due to the varying length of a sentence within each batch,  we need to pad the different sequences to the same value to facilitate the inference process. The larger the batch size is, the longer time the pad operation takes. Under the single-layer model with batch size set to 16, the inference speed is 245KB/s, which can satisfy most scenarios with strict speed requirements. 

Figure 3 shows the impact of coefficient $\lambda$. For simplicity, we only select the single-layer-based model under ratio 1\% with datasets from SIGHAN2005. The parameter is used to trade off the score between PPL and scores output by the encoder. There are two conclusions that we can draw from the picture. Firstly, under the single-layer model setting, the best coefficient for the different ratio-based models is inconsistent. Under the ratio of 1\% setting, the optimal parameter is 0.7, while with the ratio of 10\%, the optimal parameter is 0.1. Secondly, since that the coefficient $\lambda$ is used during the decoding process, it didn't take part in the training process. In reality, we can ensure the optimal $\lambda$ according to the performance on the test set to best fit the actual environment.
\begin{figure}[H]
\centering
\begin{minipage}[b]{0.49\textwidth}
\includegraphics[width=1\textwidth]{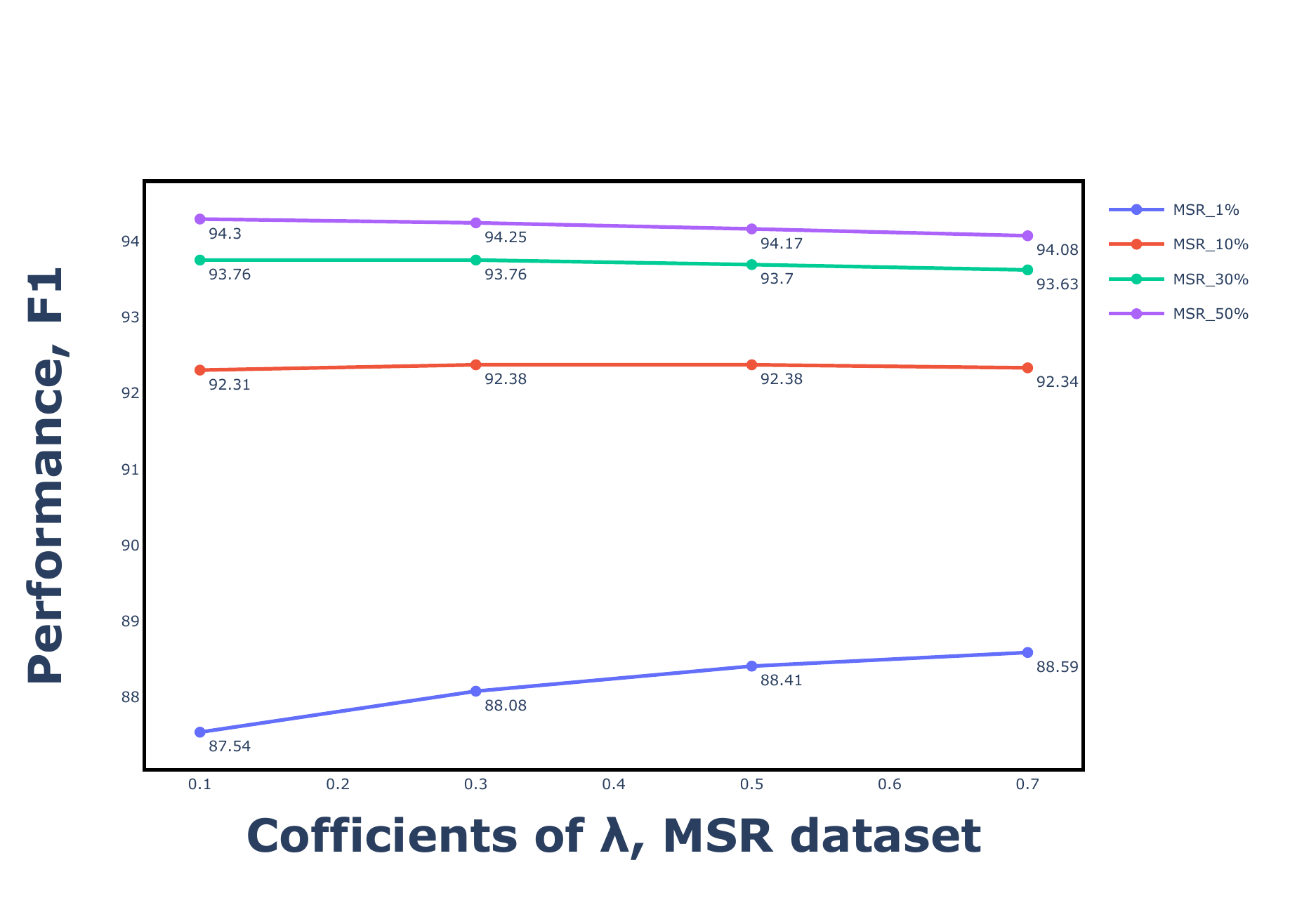}\vspace{1pt}
\end{minipage}
\hfill
\begin{minipage}[b]{0.49\textwidth}
\includegraphics[width=1\textwidth]{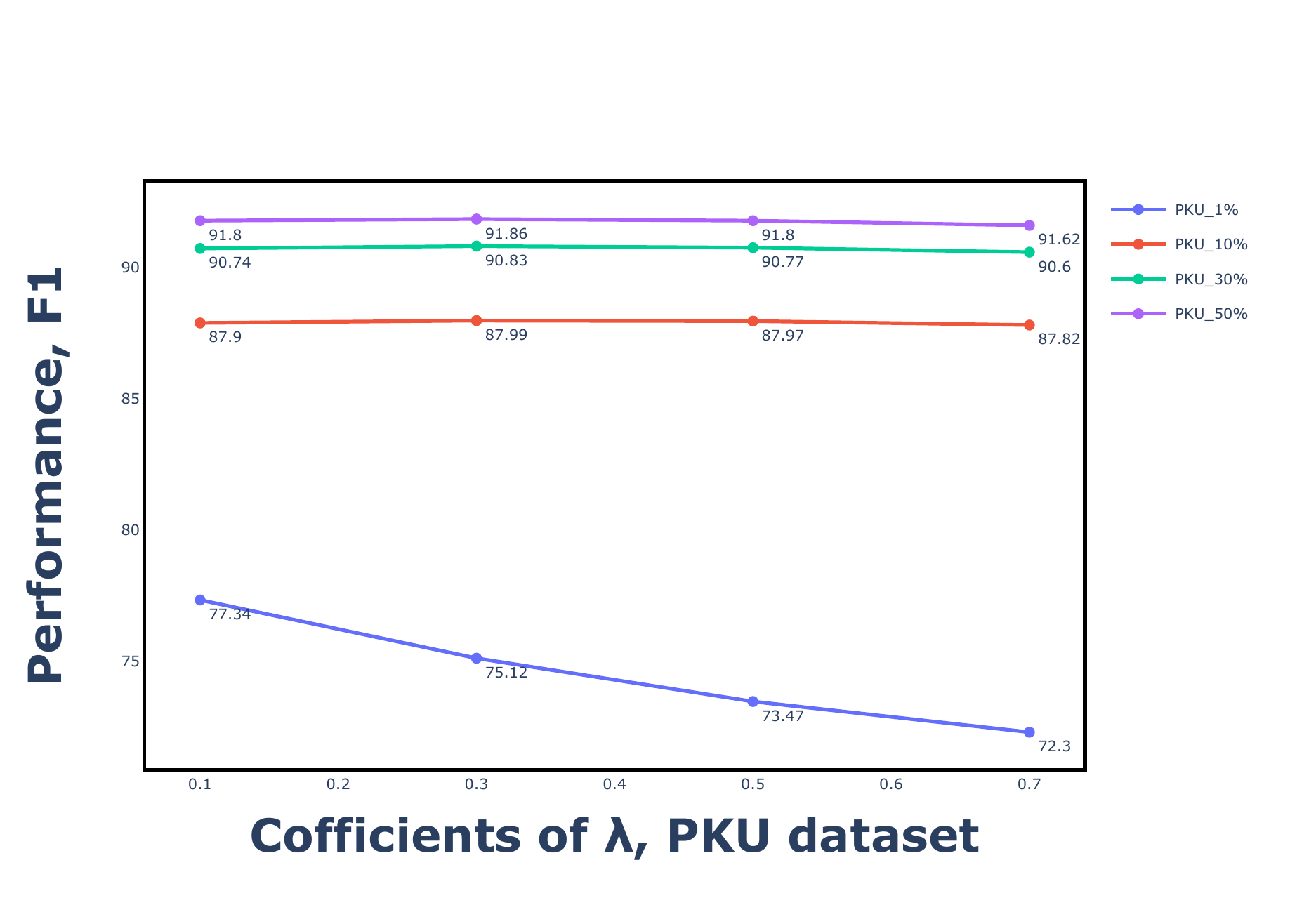}\vspace{1pt}
\end{minipage}
\vfill
\begin{minipage}[b]{0.49\textwidth}
\includegraphics[width=1\textwidth]{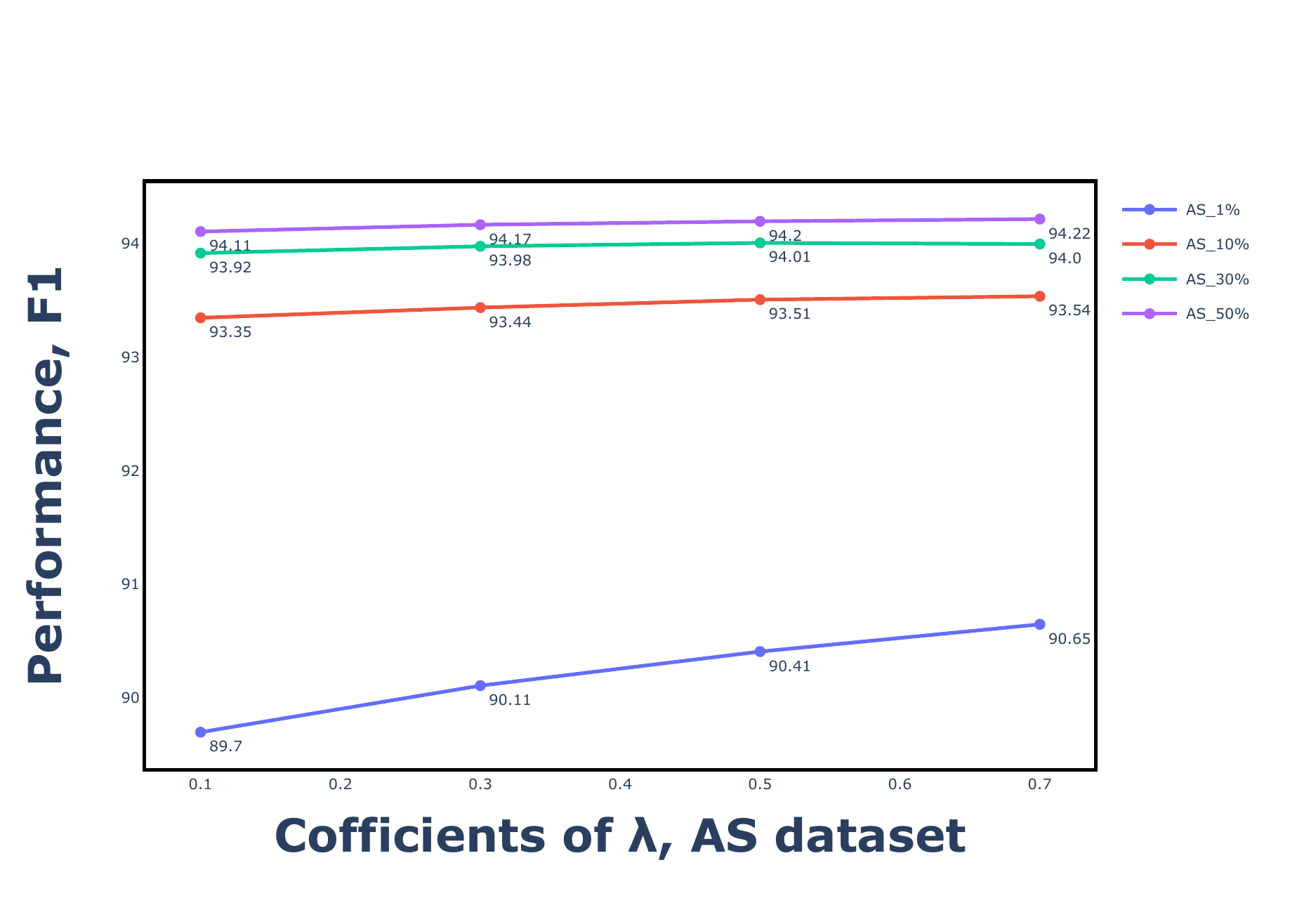}\vspace{1pt}
\end{minipage}
\hfill
\begin{minipage}[b]{0.49\textwidth}
\includegraphics[width=1\textwidth]{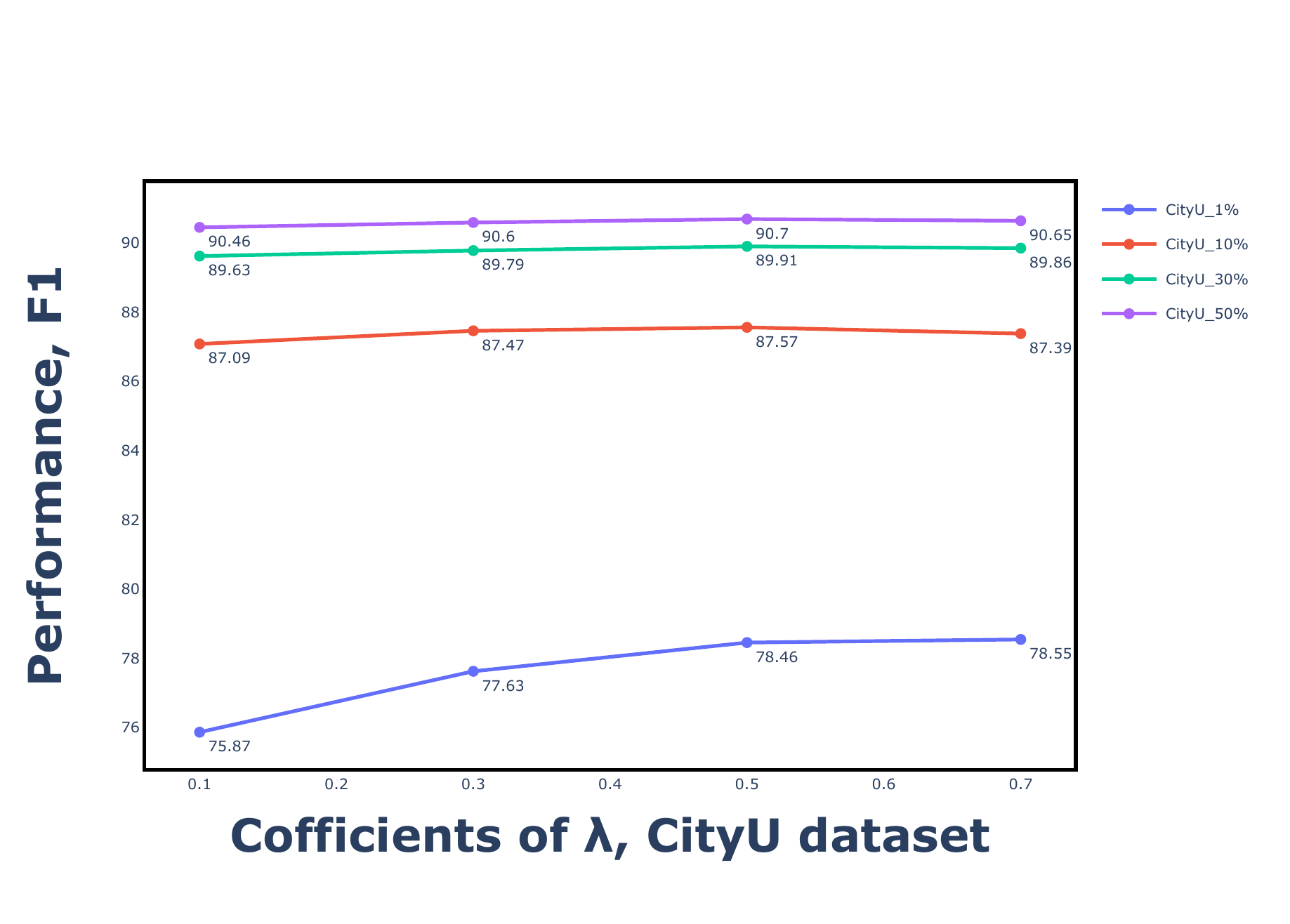}\vspace{1pt}
\end{minipage}
\caption{The effect of $\lambda$ on the performance of the model.}
\end{figure} 



\section{Conclusion}

\noindent This paper proposes an efficient CWS framework for low-resource scenarios. It is formulated to facilitate industrial use. It adopts the knowledge distillation technology to make use of the powerful pre-trained model. To best suit the practical scenario, we recommend that a distilled single-layer student model fulfill the speed requirements. To compensate for the knowledge lost during the distillation process, we propose an upgraded decode method, which introduces the n-gram language model score into the CRF, enhancing the richness of the knowledge during decoding. Experiments on diverse datasets, including several popular benchmark datasets, two publicly available datasets, and two hand-crafted datasets from real industrial scenarios, well demonstrate the effectiveness of our method.



\bibliographystyle{iclr2022_conference}
\bibliography{iclr2022_conference}


\end{document}